\documentstyle[aclap]{article}

\makeatletter

\@ifundefined{LaTeXe}{}{
   \let\normalsize\@normalsize
}

\makeatother

\newcommand{\commentout}[1] {}
\newcommand{\DATR}{{\small \sf DATR}}
\newcommand{\PATR}{{\small \sf PATR}}
\newcommand{\TAG}{{\small \sf TAG}}
\newcommand{\LTAG}{{\small \sf LTAG}}
\newcommand{\XTAG}{{\small \sf XTAG}}
\newcommand{\HPSG}{{\small \sf HPSG}}

\newcommand{\LKRL}{{\sc lkrl}}
\newcommand{\NP}{{\sc np}}
\newcommand{\PP}{{\sc pp}}
\newcommand{\VP}{{\sc vp}}
\newcommand{\Sx}{{\sc s}}
\newcommand{\Px}{{\sc p}}
\newcommand{\Vx}{{\sc v}}
\newcommand{\parentx}{{\tt parent}}
\newcommand{\leftx}{{\tt left}}
\newcommand{\rightx}{{\tt right}}
\newcommand{\topx}{{\tt top}}
\newcommand{\bottomx}{{\tt bottom}}
\newcommand{\cat}{{\tt cat}}
\newcommand{\type}{{\tt type}}

\newcommand{\anchor}{{\tt anchor}}
\newcommand{\verbx}{{\tt VERB}}
\newcommand{\verbxnp}{{\tt VERB+NP}}
\newcommand{\verbxnppp}{{\tt VERB+NP+PP}}

\title{Encoding Lexicalized Tree Adjoining Grammars with a Nonmonotonic
Inheritance Hierarchy}

\author{
Roger Evans\\
Information Technology\\
Research Institute\\
University of Brighton\\
{\tt rpe@itri.bton.ac.uk}
\And
Gerald Gazdar\\
School of Cognitive \& \\
Computing Sciences\\
University of Sussex\\
{\tt geraldg@cogs.susx.ac.uk}
\And
David Weir\\
School of Cognitive \& \\
Computing Sciences\\
University of Sussex\\
{\tt davidw@cogs.susx.ac.uk}
}

\begin{document}
\maketitle
\bibliographystyle{acl}

\begin{abstract}
This paper shows how \DATR, a widely used formal language for lexical
knowledge representation, can be used to define an \LTAG\ lexicon as
an inheritance hierarchy with internal lexical rules.  A bottom-up
featural encoding is used for \LTAG\ trees and this allows lexical
rules to be implemented as covariation constraints within feature
structures.  Such an approach eliminates the considerable redundancy
otherwise associated with an \LTAG\ lexicon.
\end{abstract}

\section{Introduction}

The Tree Adjoining Grammar~(\TAG) formalism was first introduced two
decades ago~\cite{jlt75}, and since then there has been a steady stream
of theoretical work using the formalism. But it is only more recently
that grammars of non-trivial size have been developed: Abeille, Bishop,
Cote \& Schabes~\shortcite{abcs90} describe a feature-based Lexicalized
Tree Adjoining Grammar~(\LTAG) for English which subsequently became the
basis for the grammar used in the \XTAG\ system, a wide-coverage \LTAG\
parser~\cite{dehsz94,dehs94,xtag95}. The advent of such large grammars
gives rise to questions of efficient representation, and the fully
lexicalized character of the \LTAG\ formalism suggests that recent
research into lexical representation might be a place to look for
answers (see for example Briscoe {\em et al.}\shortcite{bpc93};
Daelemans \& Gazdar\shortcite{daelgaz92}). In this paper we explore this
suggestion by showing how the lexical knowledge representation
language~(\LKRL) \DATR~\cite{eg89a,eg89b} can be used to formulate a
compact, hierarchical encoding of an \LTAG.

The issue of efficient representation for \LTAG\footnote{As with all
fully lexicalized grammar formalisms, there is really no conceptual
distinction to be drawn in \LTAG\ between the lexicon and the grammar:
the grammatical rules are just lexical properties.} is discussed by
Vijay-Shanker \& Schabes~\shortcite{vs92}, who draw attention to the
considerable redundancy inherent in \LTAG\ lexicons that are expressed
in a flat manner with no sharing of structure or properties across the
elementary trees. For example, \XTAG\ currently includes over 100,000
lexemes, each of which is associated with a family of trees (typically
around 20) drawn from a set of over 500 elementary trees.  Many of these
trees have structure in common, many of the lexemes have the same tree
families, and many of the trees within families are systematically
related in ways which other formalisms capture using transformations or
metarules.  However, the \LTAG\ formalism itself does not provide any
direct support for capturing such regularities.

Vijay-Shanker \& Schabes address this problem by introducing a
hierarchical lexicon structure with monotonic inheritance and lexical
rules, using an approach loosely based on that of
Flickinger~\shortcite{flick87} but tailored for \LTAG\ trees rather than
\HPSG\ subcategorization lists. Becker~\shortcite{becker93,becker94}
proposes a slightly different solution, combining an inheritance
component and a set of metarules\footnote{See Section~\ref{comparison}
for further discussion of these approaches.}. We share their perception
of the problem and agree that adopting a hierarchical approach provides
the best available solution to it. However, rather than creating a
hierarchical lexical formalism that is specific to the \LTAG\ problem,
we have used \DATR, an \LKRL\ that is already quite widely known and
used. From an \LTAG\ perspective, it makes sense to use an already
available \LKRL\ that was specifically designed to address these kinds
of representational issues. From a \DATR\ perspective, \LTAG\ presents
interesting problems arising from its radically lexicalist character:
{\bf all\/} grammatical relations, including unbounded dependency
constructions, are represented lexically and are thus open to lexical
generalization.

There are also several further benefits to be gained from using an
established general purpose \LKRL\ such as \DATR. First, it makes it
easier to compare the resulting \LTAG\ lexicon with those associated
with other types of lexical syntax: there are existing \DATR\ lexicon
fragments for \HPSG, \PATR\ and Word Grammar, among others. Second,
\DATR\ is not restricted to syntactic description, so one can take
advantage of existing analyses of other levels of lexical description,
such as phonology, prosody, morphology, compositional semantics and
lexical semantics\footnote{See, for example,
Bleiching~\shortcite{Bleiching92,Bleiching94}, Brown \&
Hippisley~\shortcite{Brown94}, Corbett \& Fraser~\shortcite{cf93},
Cahill~\shortcite{cahill90,cahill93}, Cahill \& Evans~\shortcite{ce90},
Fraser \& Corbett~\shortcite{Fraser94}, Gibbon~\shortcite{Gibbon92},
Kilgarriff~\shortcite{Kilgarriff93}, Kilgarriff \&
Gazdar~\shortcite{Kilgarriff95}, Reinhard \&
Gibbon~\shortcite{Reinhard91}.}. Third, one can exploit existing formal
and implementation work on the language\footnote{See, for example, Andry
{\it et al.}~\shortcite{afmty92} on compilation, Kilbury {\it et
al.}~\shortcite{Kilbury91} on coding {\sc dag}s, Duda \&
Gebhardi~\shortcite{Duda94} on dynamic querying,
Langer~\shortcite{Langer94} on reverse querying, and Barg~\shortcite{Barg94},
Light~\shortcite{Light94}, Light {\it et al.}~\shortcite{Light93} and
Kilbury {\it et al.}~\shortcite{Kilbury94} on automatic acquisition.
And there are at least a dozen different \DATR\ implementations
available, on various platforms and programming languages.}.

\section{Representing \LTAG\ trees}

\begin{figure}[thb]
\begin{centering}
\setlength{\unitlength}{0.012500in}%
\begingroup\makeatletter\ifx\SetFigFont\undefined
\def\x#1#2#3#4#5#6#7\relax{\def\x{#1#2#3#4#5#6}}%
\expandafter\x\fmtname xxxxxx\relax \def\y{splain}%
\ifx\x\y   
\gdef\SetFigFont#1#2#3{%
  \ifnum #1<17\tiny\else \ifnum #1<20\small\else
  \ifnum #1<24\normalsize\else \ifnum #1<29\large\else
  \ifnum #1<34\Large\else \ifnum #1<41\LARGE\else
     \huge\fi\fi\fi\fi\fi\fi
  \csname #3\endcsname}%
\else
\gdef\SetFigFont#1#2#3{\begingroup
  \count@#1\relax \ifnum 25<\count@\count@25\fi
  \def\x{\endgroup\@setsize\SetFigFont{#2pt}}%
  \expandafter\x
    \csname \romannumeral\the\count@ pt\expandafter\endcsname
    \csname @\romannumeral\the\count@ pt\endcsname
  \csname #3\endcsname}%
\fi
\fi\endgroup
\begin{picture}(201,193)(269,605)
\thinlines
\put(330,780){\line(-1,-1){ 40}}
\put(330,780){\line( 1,-1){ 40}}
\put(370,720){\line(-1,-1){ 40}}
\put(370,715){\line( 0,-1){ 40}}
\put(370,720){\line( 1,-1){ 40}}
\put(410,660){\line(-1,-2){ 20}}
\put(410,660){\line( 1,-2){ 20}}
\put(330,785){\makebox(0,0)[b]{\smash{\SetFigFont{10}{12.0}{rm}S}}}
\put(290,725){\makebox(0,0)[b]{\smash{\SetFigFont{10}{12.0}{rm}NP}}}
\put(370,725){\makebox(0,0)[b]{\smash{\SetFigFont{10}{12.0}{rm}VP}}}
\put(330,665){\makebox(0,0)[b]{\smash{\SetFigFont{10}{12.0}{rm}V}}}
\put(370,665){\makebox(0,0)[b]{\smash{\SetFigFont{10}{12.0}{rm}NP}}}
\put(410,665){\makebox(0,0)[b]{\smash{\SetFigFont{10}{12.0}{rm}PP}}}
\put(390,605){\makebox(0,0)[b]{\smash{\SetFigFont{10}{12.0}{rm}P}}}
\put(430,605){\makebox(0,0)[b]{\smash{\SetFigFont{10}{12.0}{rm}NP}}}
\put(300,725){\makebox(0,0)[b]{\smash{\SetFigFont{10}{12.0}{rm}$\downarrow$}}}
\put(340,665){\makebox(0,0)[b]{\smash{\SetFigFont{10}{12.0}{rm}$\diamond$}}}
\put(380,665){\makebox(0,0)[b]{\smash{\SetFigFont{10}{12.0}{rm}$\downarrow$}}}
\put(400,605){\makebox(0,0)[b]{\smash{\SetFigFont{10}{12.0}{rm}$\diamond$}}}
\put(440,605){\makebox(0,0)[b]{\smash{\SetFigFont{10}{12.0}{rm}$\downarrow$}}}
\end{picture}
\caption{An example \LTAG\ tree for {\em give}}
\label{fig1}
\end{centering}
\end{figure}
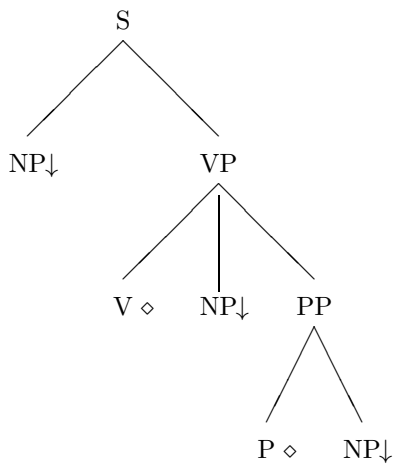

The principal unit of (syntactic) information associated with an \LTAG\
entry is a tree structure in which the tree nodes are labeled with
syntactic categories and feature information and there is at least one
leaf node labeled with a {\bf lexical\/} category (such lexical leaf
nodes are known as {\bf anchors\/}). For example, the canonical tree for
a ditransitive verb such as {\em give} is shown in figure~\ref{fig1}.
Following \LTAG\ conventions (for the time being), the node labels here
are gross syntactic category specifications to which additional featural
information may be added\footnote{In fact, \LTAG\ commonly distinguishes
two sets of features at each node (\topx\ and \bottomx), but for
simplicity we shall assume just one set in this paper.}, and are
annotated to indicate node {\bf type\/}: $\diamond$ indicates an anchor
node, and $\downarrow$ indicates a substitution node (where a fully
specified tree with a compatible root label may be
attached)\footnote{\LTAG's other tree-building operation is {\bf
adjunction}, which allows a tree-fragment to be spliced into the body of
a tree.  However, we only need to concern ourselves here with the {\bf
representation\/} of the trees involved, not with the
substitution/adjunction distinction.}.

In representing such a tree in \DATR, we do two things. First, in
keeping with the radically lexicalist character of \LTAG, we describe
the tree structure from its (lexical) anchor upwards\footnote{The tree
in figure~\ref{fig1} has more than one anchor -- in
such cases it is generally easy to decide which anchor is the most
appropriate root for the tree (here, the verb anchor).}, using a variant
of Kilbury's~\shortcite{Kilbury90} bottom-up encoding of trees.  In this
encoding, a tree is described relative to a particular distinguished
leaf node (here the anchor node), using binary relations \parentx,
\leftx\ and \rightx, relating the node to the subtrees associated with
its parent, and immediate-left and -right sisters, encoded in the same
way. Second, we embed the resulting tree structure (i.e., the node
relations and type information) in the feature structure, so that the
tree relations (\leftx, \rightx\ and \parentx) become features.  The
obvious analogy here is the use of {\tt first}/{\tt rest} features to
encode subcategorisation lists in frameworks like \HPSG.

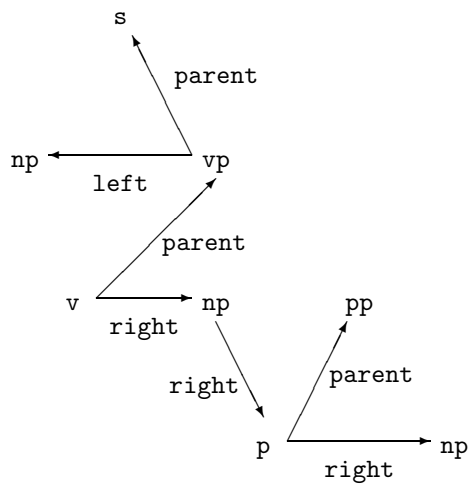
\begin{figure}[thb]
\begin{centering}
\setlength{\unitlength}{0.012500in}%
\begingroup\makeatletter\ifx\SetFigFont\undefined
\def\x#1#2#3#4#5#6#7\relax{\def\x{#1#2#3#4#5#6}}%
\expandafter\x\fmtname xxxxxx\relax \def\y{splain}%
\ifx\x\y   
\gdef\SetFigFont#1#2#3{%
  \ifnum #1<17\tiny\else \ifnum #1<20\small\else
  \ifnum #1<24\normalsize\else \ifnum #1<29\large\else
  \ifnum #1<34\Large\else \ifnum #1<41\LARGE\else
     \huge\fi\fi\fi\fi\fi\fi
  \csname #3\endcsname}%
\else
\gdef\SetFigFont#1#2#3{\begingroup
  \count@#1\relax \ifnum 25<\count@\count@25\fi
  \def\x{\endgroup\@setsize\SetFigFont{#2pt}}%
  \expandafter\x
    \csname \romannumeral\the\count@ pt\expandafter\endcsname
    \csname @\romannumeral\the\count@ pt\endcsname
  \csname #3\endcsname}%
\fi
\fi\endgroup
\begin{picture}(211,203)(294,595)
\thinlines
\put(340,670){\vector( 1, 0){ 40}}
\put(380,730){\vector(-1, 2){ 25}}
\put(380,730){\vector(-1, 0){ 60}}
\put(390,660){\vector( 1,-2){ 20}}
\put(420,610){\vector( 1, 0){ 60}}
\put(420,610){\vector( 1, 2){ 25}}
\put(340,670){\vector( 1, 1){ 50}}
\put(350,785){\makebox(0,0)[b]{\smash{\SetFigFont{10}{12.0}{rm}{\tt s}}}}
\put(390,760){\makebox(0,0)[b]{\smash{\SetFigFont{10}{12.0}{rm}{\tt parent}}}}
\put(310,725){\makebox(0,0)[b]{\smash{\SetFigFont{10}{12.0}{rm}{\tt np}}}}
\put(350,715){\makebox(0,0)[b]{\smash{\SetFigFont{10}{12.0}{rm}{\tt left}}}}
\put(390,725){\makebox(0,0)[b]{\smash{\SetFigFont{10}{12.0}{rm}{\tt vp}}}}
\put(385,690){\makebox(0,0)[b]{\smash{\SetFigFont{10}{12.0}{rm}{\tt parent}}}}
\put(330,665){\makebox(0,0)[b]{\smash{\SetFigFont{10}{12.0}{rm}{\tt v}}}}
\put(360,655){\makebox(0,0)[b]{\smash{\SetFigFont{10}{12.0}{rm}{\tt right}}}}
\put(390,665){\makebox(0,0)[b]{\smash{\SetFigFont{10}{12.0}{rm}{\tt np}}}}
\put(410,605){\makebox(0,0)[b]{\smash{\SetFigFont{10}{12.0}{rm}{\tt p}}}}
\put(450,665){\makebox(0,0)[b]{\smash{\SetFigFont{10}{12.0}{rm}{\tt pp}}}}
\put(455,635){\makebox(0,0)[b]{\smash{\SetFigFont{10}{12.0}{rm}{\tt parent}}}}
\put(450,595){\makebox(0,0)[b]{\smash{\SetFigFont{10}{12.0}{rm}{\tt right}}}}
\put(490,605){\makebox(0,0)[b]{\smash{\SetFigFont{10}{12.0}{rm}{\tt np}}}}
\put(385,630){\makebox(0,0)[b]{\smash{\SetFigFont{10}{12.0}{rm}{\tt right}}}}
\end{picture}
\caption{Bottom-up encoding for {\tt Give}}
\label{fig2}
\end{centering}
\end{figure}

Thus the syntactic feature information directly associated with the
entry for {\em give} relates to the label for the \Vx\ node (for example,
the value of its \cat\ feature is {\tt v}, the value of \type\ is \anchor),
while specifications of subfeatures of \parentx\ relate to the label of
the \VP\ node. A simple bottom-up \DATR\ representation for the whole tree
(apart from the node type information) follows:
{\small
\begin{verbatim}
Give:
    <cat> = v
    <parent cat> = vp
    <parent left cat> = np
    <parent parent cat> = s
    <right cat> = np
    <right right cat> = p
    <right right parent cat> = pp
    <right right right cat> = np.
\end{verbatim}
} This says that {\tt Give} is a verb, with \VP\ as its parent, an \Sx\
as its grandparent and an \NP\ to the left of its parent. It also has an
\NP\ to its right, and a tree rooted in a \Px\ to the right of that,
with a \PP\ parent and \NP\ right sister.  The implied bottom-up tree
structure is shown graphically in figure~\ref{fig2}.  Here the nodes are
laid out just as in figure~\ref{fig1}, but related via \parentx, \leftx\
and \rightx\ links, rather than the more usual (implicitly ordered)
daughter links. Notice in particular that the \rightx\ link from the
object noun-phrase node points to the {\em preposition\/} node, not its
phrasal parent -- this whole subtree is itself encoded
bottom-up. Nevertheless, the full tree structure is completely and
accurately represented by this encoding.

Once we adopt this representational strategy, writing an \LTAG\
lexicon in \DATR\ becomes similar to writing any other type
of lexicalist grammar's lexicon in an inheritance-based \LKRL. In \HPSG,
for example, the subcategorisation frames are coded as lists of
categories, whilst in \LTAG\ they are coded as trees. But, in both cases,
the problem is one of concisely describing feature structures
associated with lexical entries and relationships between lexical
entries. The same kinds of generalization arise
and the same techniques are applicable.
Of course, the presence of complete trees and the fully
lexicalized approach provide scope for capturing generalizations
lexically that are not available to approaches that only identify
parent and sibling nodes, say, in the lexical entries.

\section{Encoding lexical entries}
\label{canonical-lex}

\begin{figure}[thb]
\begin{centering}
\setlength{\unitlength}{0.012500in}%
\begingroup\makeatletter\ifx\SetFigFont\undefined
\def\x#1#2#3#4#5#6#7\relax{\def\x{#1#2#3#4#5#6}}%
\expandafter\x\fmtname xxxxxx\relax \def\y{splain}%
\ifx\x\y   
\gdef\SetFigFont#1#2#3{%
  \ifnum #1<17\tiny\else \ifnum #1<20\small\else
  \ifnum #1<24\normalsize\else \ifnum #1<29\large\else
  \ifnum #1<34\Large\else \ifnum #1<41\LARGE\else
     \huge\fi\fi\fi\fi\fi\fi
  \csname #3\endcsname}%
\else
\gdef\SetFigFont#1#2#3{\begingroup
  \count@#1\relax \ifnum 25<\count@\count@25\fi
  \def\x{\endgroup\@setsize\SetFigFont{#2pt}}%
  \expandafter\x
    \csname \romannumeral\the\count@ pt\expandafter\endcsname
    \csname @\romannumeral\the\count@ pt\endcsname
  \csname #3\endcsname}%
\fi
\fi\endgroup
\begin{picture}(222,193)(312,605)
\thinlines
\put(410,720){\line(-2,-1){ 80}}
\put(410,720){\line( 0,-1){ 40}}
\put(410,720){\line( 2,-1){ 80}}
\put(410,660){\line( 0,-1){ 40}}
\put(490,660){\line( 0,-1){ 40}}
\put(370,780){\line(-1,-1){ 40}}
\put(370,780){\line( 1,-1){ 40}}
\put(410,725){\makebox(0,0)[b]{\smash{\SetFigFont{10}{12.0}{rm}{\tt VERB+NP}}}}
\put(330,665){\makebox(0,0)[b]{\smash{\SetFigFont{10}{12.0}{rm}{\tt Eat}}}}
\put(410,665){\makebox(0,0)[b]{\smash{\SetFigFont{10}{12.0}{rm}{\tt
VERB+NP+PP}}}}
\put(490,665){\makebox(0,0)[b]{\smash{\SetFigFont{10}{12.0}{rm}{\tt
VERB+NP+NP}}}}
\put(410,605){\makebox(0,0)[b]{\smash{\SetFigFont{10}{12.0}{rm}{\tt Give}}}}
\put(490,605){\makebox(0,0)[b]{\smash{\SetFigFont{10}{12.0}{rm}{\tt Spare}}}}
\put(370,785){\makebox(0,0)[b]{\smash{\SetFigFont{10}{12.0}{rm}{\tt VERB}}}}
\put(330,725){\makebox(0,0)[b]{\smash{\SetFigFont{10}{12.0}{rm}{\tt Die}}}}
\end{picture}
\caption{The principal lexical hierarchy}
\label{fig3}
\end{centering}
\end{figure}
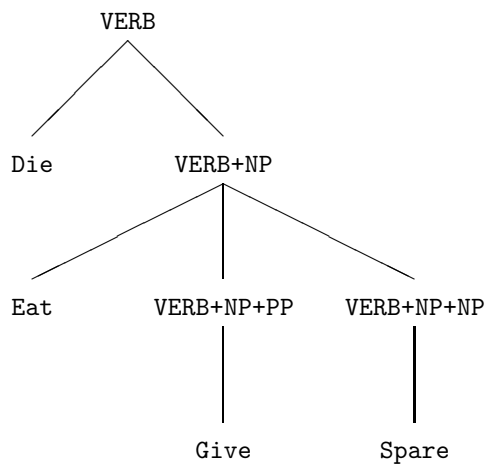

Following conventional models of lexicon organisation, we would expect
{\tt Give} to have a minimal syntactic specification itself, since
syntactically it is
a completely regular ditransitive verb. In fact {\bf none\/} of the
information introduced so far is specific to {\tt Give}. So rather than
providing a completely explicit \DATR\ definition for {\tt Give}, as we
did above, a more plausible account uses an inheritance hierarchy
defining abstract intransitive, transitive and ditransitive verbs to
support {\tt Give} (among others), as shown in figure~\ref{fig3}.

This basic organisational structure can be expressed as the following
\DATR\ fragment\footnote{To gain the intuitive sense of this fragment,
read a line such as {\tt <> == VERB} as ``inherit everything from
the definition of \verbx '', and a line such as {\tt <parent> ==
PPTREE:<>} as ``inherit the \parentx\ subtree from the definition
of {\tt PPTREE}''. Inheritance in \DATR\ is always by default --
locally defined feature specifications take priority over inherited
ones.}:
{\small
\begin{verbatim}
VERB:
    <> == TREENODE
    <cat> == v
    <type> == anchor
    <parent> == VPTREE:<>.

VERB+NP:
    <> == VERB
    <right> == NPCOMP:<>.

VERB+NP+PP:
    <> == VERB+NP
    <right right> == PTREE:<>
    <right right root> == to.

VERB+NP+NP:
    <> == VERB+NP
    <right right> == NPCOMP:<>.

Die:
    <> == VERB
    <root> == die.

Eat:
    <> == VERB+NP
    <root> == eat.

Give:
    <> == VERB+NP+PP
    <root> == give.

Spare:
    <> == VERB+NP+NP
    <root> == spare.
\end{verbatim}
}
Ignoring for the moment the references to {\tt TREENODE, VPTREE, NPCOMP}
and {\tt PTREE} (which we shall define shortly), we see that \verbx\
defines basic features for all verb entries (and can be used directly
for intransitives such as {\tt Die}), \verbxnp\ inherits from \verbx\
but adds an \NP\ complement to the right of the verb (for transitives),
\verbxnppp\ inherits from \verbxnp\ but adds a further PP complement
and so on. Entries for regular verb lexemes are then minimal --
syntactically they just inherit {\bf everything\/} from the abstract
definitions.

This \DATR\ fragment is incomplete, because it neglects to define the
internal structure of the {\tt TREENODE} and the various
subtree nodes in the lexical hierarchy.
Each such node is a
description of an \LTAG\ tree at some degree of
abstraction\footnote{Even the lexeme nodes are abstract -- individual
word forms might be represented by further more specific nodes attached
below the lexemes in the hierarchy.}. The following \DATR\ statements
complete the fragment, by providing definitions for this internal
structure:
{\small
\begin{verbatim}
TREENODE:
    <> == undef
    <type> == internal.

STREE:
    <> == TREENODE
    <cat> == s.

VPTREE:
    <> == TREENODE
    <cat> == vp
    <parent> == STREE:<>
    <left> == NPCOMP:<>.

NPCOMP:
    <> == TREENODE
    <cat> == np
    <type> == substitution.

PPTREE:
    <> == TREENODE
    <cat> == pp.

PTREE:
    <> == TREENODE
    <cat> ==  p
    <type> == anchor
    <parent> == PPTREE:<>
\end{verbatim}
}
Here, {\tt TREENODE} represents an abstract node in an \LTAG\ tree and
provides a (default) \type\ of {\tt internal}. Notice that \verbx\ is
itself a {\tt TREENODE} (but with the nondefault type \anchor),
and the other definitions here define the remaining tree nodes that
arise in our small lexicon: {\tt VPTREE} is the node for \verbx's
parent, {\tt STREE} for \verbx's grandparent, {\tt NPCOMP} defines the
structure needed for \NP\ complement substitution nodes, etc.\footnote{
Our example makes much use of multiple inheritance (thus, for example,
{\tt VPTREE} inherits from {\tt TREENODE}, {\tt STREE} and {\tt NPCOMP})
but all such multiple inheritance is orthogonal in \DATR: no path
can inherit from more than one node.}

Taken together, these definitions provide a specification for {\tt Give}
just as we had it before, but with the addition of \type\ and {\tt root}
features. They also support some other verbs too, and it should be
clear that the basic technique extends readily to a wide range of other
verbs and other parts of speech. Also, although the trees we have
described are all {\bf initial} trees (in \LTAG\ terminology), we can
describe {\bf auxiliary} trees, which include a leaf node of type
{\tt foot} just as easily. A simple example is provided by the following
definition for auxiliary verbs:
{\small
\begin{verbatim}
AUXVERB:
    <> == TREENODE
    <cat> == v
    <type> == anchor
    <parent cat> == vp
    <right cat> == vp
    <right type> == foot.
\end{verbatim}
}

\section{Lexical rules}

Having established a basic structure for our \LTAG\ lexicon, we now turn
our attention towards capturing other kinds of relationship among trees.
We noted above that lexical entries are actually associated with
{\bf tree families\/}, and that these group together trees that
are related to each other. Thus in the same family as a standard
ditransitive verb, we might find the full passive, the agentless
passive, the dative alternation, the various relative clauses, and so
forth. It is clear that these families correspond closely to the outputs
of transformations or metarules in other frameworks, but the \XTAG\
system currently has no formal component for describing the
relationships among families nor mechanisms for generating them. And so
far we have said nothing about them either -- we have only characterized
single trees.

However, \LTAG 's large domain of locality means that {\bf all} such
relationships can be viewed as directly lexical, and thus expressible
by lexical rules.  In fact we can go further than this: because we
have embedded the domain of these lexical rules, namely the \LTAG\
tree structures, within the feature structures, we can view such
lexical rules as covariation constraints within feature structures,
in much the same way that the covariation of, say, syntactic and
morphological form is treated. In particular, we can use the
mechanisms that \DATR\ already provides for feature covariation,
rather than having to invoke in addition some special purpose lexical
rule machinery.

We consider six construction types found in the \XTAG\
grammar: passive, dative, subject-auxiliary inversion, {\em
wh}-questions, relative clauses and topicalisation. Our basic
approach to each of these is the same. Lexical rules are specified by
defining a derived {\tt output} tree structure in terms of an {\tt
input} tree structure, where each of these structures is a set of
feature specifications of the sort defined above. Each lexical rule
has a name, and the input and output tree structures for rule {\tt
foo} are referenced by prefixing feature paths of the sort given
above with {\tt <input foo ..>} or {\tt <output foo ..>}. So for
example, the category of the parent tree node of the output of the
passive rule might be referenced as {\tt <output passive parent
cat>}. We define a very general default, stating that the {\tt
output} is the same as the {\tt input}, so that lexical relationships
need only concern themselves with components they modify.  This
approach to formulating lexical rules in \DATR\ is quite general and
in no way restricted to \LTAG: it can be readily adapted for
application in the context of any feature-based lexicalist grammar
formalism.

Using this approach, the dative lexical rule can be given a
minimalist implementation by the addition of the following single
line to \verbxnppp, defined above.
{\small
\begin{verbatim}
VERB+NP+PP:
    <output dative right right> == NPCOMP:<>.
\end{verbatim}
}
This causes the second complement to a ditransitive verb in the dative
alternation to be an \NP, rather than a \PP\ as in the unmodified case.
Subject-auxiliary inversion can be achieved similarly by just specifying
the {\tt output} tree structure without reference to the {\tt input}
structure (note the addition here of a {\tt form} feature specifying
verb form):
{\small
\begin{verbatim}
AUXVERB:
    <output auxinv form> == finite-inv
    <output auxinv parent cat> == s
    <output auxinv right cat> == s.
\end{verbatim}
}

Passive is slightly more complex, in that it has to modify the given
{\tt input} tree structure rather than simply overwriting part of it.
The definitions for passive occur at the \verbxnp\ node, since by
default, any transitive or subclass of transitive has a passive form.
Individual transitive verbs, or whole subclasses, can override
this default, leaving their passive tree structure undefined if
required. For agentless passives, the necessary additions to the
\verbxnp\ node are as follows\footnote{Oversimplifying slightly, the
double quotes in {\tt "<input passive right right>"} mean that that
\DATR\ path will not be evaluated locally (i.e., at the {\tt VERB+NP}
node), but rather at the relevant lexeme node (e.g., {\tt Eat} or {\tt
Give}).}:
{\small
\begin{verbatim}
VERB+NP:
    <output passive form> == passive
    <output passive right> ==
            "<input passive right right>".
\end{verbatim}
}
Here, the first line stipulates the form of the verb in the output tree
to be passive, while the second line redefines the complement structure:
the {\tt output} of passive has as its first complement the second
complement of its {\tt input}, thereby discarding the first complement
of its {\tt input}. Since complements are daisy-chained, all the
others move up too.

{\em Wh}-questions, relative clauses and topicalisation are slightly
different, in that the application of the lexical rule causes structure
to be added to the top of the tree (above the \Sx\ node). Although
these constructions involve unbounded dependencies, the unboundedness is
taken care of by the \LTAG\ adjunction mechanism: for lexical purposes
the dependency is local. Since the relevant lexical rules can apply to
sentences that contain any kind of verb, they need to be stated at the
\verbx\ node. Thus, for example, topicalisation and {\em wh}-questions
can be defined as follows:
{\small
\begin{verbatim}
VERB:
    <output topic parent parent parent cat>
                                           == s
    <output topic parent parent left cat> == np
    <output topic parent parent left form>
                                      == normal
    <output whq> == "<output topic>"
    <output whq parent parent left form> == wh.
\end{verbatim}
}
Here an additional \NP\ and \Sx\ are attached above the original \Sx\ node
to create a topicalised structure. The {\em wh}-rule inherits from the
topicalisation rule, changing just one thing: the form of the new
\NP\ is marked as {\tt wh}, rather than as {\tt normal}.
In the full fragment\footnote{The full version of this \DATR\ fragment includes
all
the components discussed above in a single coherent, but slightly more
complex account. It is available on request from the authors.},
the \NP\ added by these rules is also syntactically cross-referenced to
a specific \NP\ marked as null in the {\tt input} tree.
However, space does not permit presentation or discussion of the
\DATR\ code that achieves this here.

\section{Applying lexical rules}

As explained above, each lexical rule is defined to operate on its own
notion of an {\tt input} and produce its own {\tt output}. In order for
the rules to have an effect, the various {\tt input} and {\tt output}
paths have to be linked together using inheritance, creating a chain of
inheritances between the base, that is, the canonical definitions
we introduced in section \ref{canonical-lex}, and {\tt surface} tree
structures of the lexical entry. For example, to `apply' the dative rule
to our {\tt Give} definition, we could construct a definition such as
this:
{\small
\begin{verbatim}
Give-dat:
    <> == Give
    <input dative> == <>
    <surface> == <output dative>.
\end{verbatim}
}
Values for paths prefixed with {\tt surface} inherit from the output of
the dative rule. The input of the dative rule inherits from the base
(unprefixed) case, which inherits from {\tt Give}. The dative rule
definition (just the one line introduced above, plus the default that
output inherits from input) thus mediates between {\tt Give} and the
surface of {\tt Give-dat}. This chain can be
extended by inserting additional inheritance specifications (such as
passive). Note that {\tt surface} defaults to the base
case, so all entries have a {\tt surface} defined.

However, in our full fragment, additional support is provided to achieve
and constrain this rule chaining. Word definitions include boolean
features indicating which rules to apply, and the presence of these
features trigger inheritance between appropriate {\tt input} and {\tt
output} paths and the base and {\tt surface} specifications at the
ends of the chain. For example, {\tt Word1} is an alternative way of
specifying the dative alternant of {\tt Give}, but results in
inheritance linking equivalent to that found in {\tt Give-dat} above:
{\small
\begin{verbatim}
Word1:
    <> == Give
    <alt dative> == true.
\end{verbatim}
}
More interestingly, {\tt Word2} properly describes a {\em
wh}-question based on the agentless passive of the dative of {\tt Give}.
{\small \begin{verbatim} Word2:
    <> == Give
    <alt whq> == true
    <alt dative> == true
    <alt passive> == true.
    <parent left form> == null
\end{verbatim}
}
Notice here the final line of {\tt Word2} which specifies the location
of the `extracted' \NP\ (the subject, in this case), by marking it
as null. As noted above, the full version of the {\tt whq} lexical
rule uses this to specify a cross-reference relationship between the
{\em wh}-\NP\ and the null \NP.

We can, if we wish, encode constraints on the applicability of
rules in the mapping from boolean flags to actual inheritance
specifications. Thus, for example, {\tt whq, rel,} and {\tt topic} are
mutually exclusive.  If such constraints are violated, then no value for
{\tt surface} gets defined.
Thus {\tt Word3} improperly attempts topicalisation in addition to {\em
wh}-question formation, and, as a result, will fail to define a {\tt
surface} tree structure at all:
{\small \begin{verbatim} Word3:
    <> == Give
    <alt whq> == true
    <alt topic> == true
    <alt dative> == true
    <alt passive> == true
    <parent left form> == null.
\end{verbatim}
}

This approach to lexical rules allows them to be specified at the
appropriate point in the lexical hierarchy, but overridden or
modified in subclasses or lexemes as appropriate. It also allows
default generalisation over the lexical rules themselves, and control
over their application. The last section showed how the {\tt whq}
lexical rule could be built by a single minor addition to that for
topicalisation. However, it is worth noting that, in common with
other \DATR\ specifications, the lexical rules presented here are
{\bf rule instances\/} which can only be applied once to any given
lexeme -- multiple application could be supported, by making multiple
instances inherit from some common rule specification, but in our
current treatment such instances would require different rule names.

\section{Comparison with related work}
\label{comparison}

As noted above, Vijay-Shanker \& Schabes~\shortcite{vs92} have also
proposed an inheritance-based approach to this problem. They use
monotonic inheritance to build up partial descriptions of trees: each
description is a finite set of dominance, immediate dominance and linear
precedence statements about tree nodes in a tree description language
developed by Rogers \& Vijay-Shanker~\shortcite{Rogers92}, and category
information is located in the node labels.

This differs from our approach in a number of ways. First, our use of
nonmonotonic inheritance allows us to manipulate total instead of
partial descriptions of trees. The abstract verb class in the Vijay-Shanker
\& Schabes account subsumes both intransitive and transitive verb classes
but is not identical to either -- a minimal-satisfying-model step is
required to map partial tree descriptions into actual trees.
In our analysis,
\verbx\ {\bf is} the intransitive verb class, with complements
specifically marked as undefined: thus {\tt VERB:<right> == undef} is
inherited from {\tt TREENODE} and \verbxnp\ just
overrides this complement specification to add an \NP\ complement.
Second, we describe trees using only local tree relations (between
adjacent nodes in the tree), while Vijay-Shanker \& Schabes also use a
nonlocal dominance relation.

Both these properties are crucial to our embedding of the tree structure
in the feature structure. We want the category information at each tree
node to be partial in the conventional sense, so that in actual use such
categories can be extended (by unification or whatever). So the feature
structures that we associate with lexical entries must be viewed as
partial. But we do {\bf not} want the tree structure to be extendible in
the same way: we do not want an intransitive verb to be applicable in a
transitive context, by unifying in a complement \NP. So the tree
structures we define must be total descriptions\protect\footnote {Note
that simplified fragment presented here does not get this right. It
makes all feature specifications total descriptions. To correct this we
would need to change {\tt TREENODE} so that only the values of {\tt
<right>}, {\tt <left>} and {\tt <parent>} default to {\tt undef}.}.  And
of course, our use of only local relations allows a direct mapping from
tree structure to feature path, which would not be possible at all if
nonlocal relations were present.

So while these differences may seem small, they allow us to take this
significant representational step -- significant because it is the tree
structure embedding that allows us to view lexical rules as feature
covariation constraints. The result is that while Vijay-Shanker \&
Schabes use a tree description language, a category description
language and a further formalism for lexical rules, we can capture
everything in one framework all of whose components (nonmonotonicity,
covariation constraint handling, etc.) have already been independently
motivated for other aspects of lexical description\footnote{As in the work
cited in footnote 3, above.}.

Becker's recent work~\shortcite{becker93,becker94} is also directed at
exactly the problem we address in the present paper.  Like him, we have
employed an inheritance hierarchy.  And, like him, we have employed a
set of lexical rules (corresponding to his metarules).  The key
differences between our account and his are (i) that we have been able
to use an existing lexical knowledge representation language, rather
than designing a formal system that is specific to \LTAG , and (ii) that
we have expressed our lexical rules in exactly the same language as that
we have used to define the hierarchy, rather than invoking two quite
different formal systems.

Becker's sharp distinction between his metarules and his hierarchy
gives rise to some problems that our approach avoids.  Firstly, he
notes that his metarules are subject to lexical exceptions and
proposes to deal with these by stating ``for each entry in the
(syntactic) lexicon .. which metarules are applicable for this entry''
(1993,126).  We have no need to carry over this use of (meta)rule
features since, in our account, lexical rules are not distinct from
any other kind of property in the inheritance hierarchy.  They can be
stated at the most inclusive relevant node and can then be overridden
at the exceptional descendant nodes.  Nothing specific needs to be
said about the nonexceptional nodes.

Secondly, his metarules may themselves be more or less similar to
each other and he suggests (1994,11) that these similarities could
be captured if the metarules were also to be organized in a
hierarchy.  However, our approach allows us to deal with any such
similarities in the main lexical hierarchy itself\footnote{As
illustrated by the way in which the {\tt whq} lexical rule inherits
from that for topicalisation in the example given above.} rather than by
setting up a separate hierarchical component just for metarules
(which appears to be what Becker has in mind).

Thirdly, as he himself notes (1993,128), because his metarules map
from elementary trees that are in the inheritance hierarchy to
elementary trees that are outside it, most of the elementary trees
actually used are not directly connected to the hierarchy (although
their derived status with respect to it can be reconstructed).  Our
approach keeps all elementary trees, whether or not they have been
partly defined by a lexical rule, entirely within the lexical
hierarchy.

In fact, Becker himself considers the possibility of capturing all
the significant generalizations by using just one of the two
mechanisms that he proposes: ``one might want to reconsider the usage
of one mechanism for phenomena in both dimensions'' (1993,135).
But, as he goes on to point out, his existing type of inheritance
network is not up to taking on the task performed by his metarules
because the former is monotonic whilst his metarules are not.
However, he does suggest a way in which the hierarchy could be
completely replaced by metarules but argues against adopting it
(1993,136).

As will be apparent from the earlier sections of this paper, we
believe that Becker's insights about the organization of an \LTAG\
lexicon can be better expressed if the metarule component is replaced
by an encoding of (largely equivalent) lexical rules that are an
integral part of a nonmonotonic inheritance hierarchy that stands as
a description of {\bf all} the elementary trees.

\section*{Acknowledgements}

A precursor of this paper was presented at the September 1994 {\sf TAG+}
Workshop in Paris.  We thank the referees for that event and the ACL-95
referees for a number of helpful comments.  We are also grateful to
Aravind Joshi, Bill Keller, Owen Rambow K. Vijay-Shanker and The XTAG
Group.  This research was partly supported by grants to Evans from
SERC/EPSRC (UK) and to Gazdar from ESRC (UK).

{\small

}

\end{document}